\documentclass[10pt,twocolumn]{article}

\usepackage{times}
\usepackage{mathptmx}

\usepackage{amsmath,amssymb}
\usepackage{graphicx}
\usepackage{booktabs}
\usepackage{url}
\usepackage[colorlinks=true, citecolor=blue, linkcolor=black, urlcolor=blue]{hyperref}
\usepackage[top=2cm, bottom=2cm, left=1.5cm, right=1.5cm, columnsep=1.5cm]{geometry}
\usepackage{microtype}
\usepackage{float}
\usepackage{placeins}
\usepackage{tikz}
\usepackage{mathtools}
\usepackage[round]{natbib}
\usepackage{cuted}
\usepackage{multirow}
\title{\textbf{LFC-DA: Logical Formula-Controlled Data Augmentation for Enhanced Logical Reasoning}}

\author{Shenghao Li\textsuperscript{1}}
\date{
    \small
    \textsuperscript{1}Guangzhou University, Guangzhou, China \\
    \small \texttt{lishenghao@e.gzhu.edu.cn}
}

\begin{document}

\maketitle

\begin{center}
\textbf{Abstract}
\end{center}
\noindent
For complex logical data augmentation, heavy reliance on human annotation is costly, whereas direct generation with large language models yields uninterpretable and logically homogeneous examples. To address this, we present LFC-DA, a symbolic-logic-controlled pipeline: logical text is first mapped to propositional expressions, a compact rule library is compiled, and a bounded state-space search systematically discovers valid formulas that are then verbalized back into natural-language questions, ensuring both diversity and logical rigor under propositional logic. Experiments on ReClor and LogiQA show significant improvements in the logical-reasoning accuracy of pretrained models, confirming the effectiveness of LFC-DA for LLM-guided logical data augmentation.

\section{Introduction}
In recent years, owing to the powerful generation and comprehension capabilities of large language models (LLMs), they have demonstrated exceptional performance in tasks such as text generation and question answering\citep{zhao2023survey} and have been widely deployed in vertical domains like education and the legal sector.However, the systematic deficiencies in logical reasoning exhibited by existing pre-trained models severely constrain their reliable deployment in real-world high-risk scenarios. Addressing this issue through manual annotation is time-consuming, labor-intensive, and difficult to scale\citep{ouyang2022training}.Although leveraging LLMs themselves for logical data augmentation serves as an efficient alternative, it faces two fundamental challenges: first, the lack of interpretability in the generation process makes it difficult for models to guarantee the logical correctness of the generated outcomes\citep{huang2023large}.second, the generated results suffer from limited diversity in logical forms, as models tend to produce superficial linguistic variations rather than substantive variations in logical structure \citep{feng2021survey}, resulting in augmented data that fails to support models in learning deep reasoning patterns.

To address the aforementioned challenges, this paper proposes LFC-DA, a formally-logic-controlled data augmentation method. The core design philosophy of LFC-DA is to establish a controllable pipeline of "Formalization-Exploration-Instantiation", as illustrated in Figure 1. Specifically, natural language statements are first formalized into propositional logic formulas to establish a logical starting point. Subsequently, novel formulas within the same logical system are systematically explored. Finally, large language models are driven to accurately instantiate these logical forms back into natural text. The generated text can then serve as training data for discriminative pre-trained models.

The main contributions of this paper are as follows:
\begin{enumerate}
    \item We propose a symbolic logic-based controllable data augmentation framework (LFC-DA). This framework constructs a formal logical rule base to constrain data generation within an interpretable symbolic logic framework, addressing the issues of low logical fidelity in augmented data and the lack of interpretability in the generation process.
    
    \item We design a state-space search algorithm based on backtracking. This algorithm can systematically explore complex logical expressions within the logic space defined by the rule base, ensuring both structural diversity and formal rigor of the generated data under propositional logic.
    
    \item Experimental results demonstrate that the data generated by our method can significantly enhance the reasoning performance of pre-trained models (e.g., RoBERTa\citep{liu2019robustly}), effectively validating its efficacy and generalization capability.
\end{enumerate}

\begin{figure*}[htbp]
    \centering
    \fbox{\includegraphics[width=0.9\textwidth]{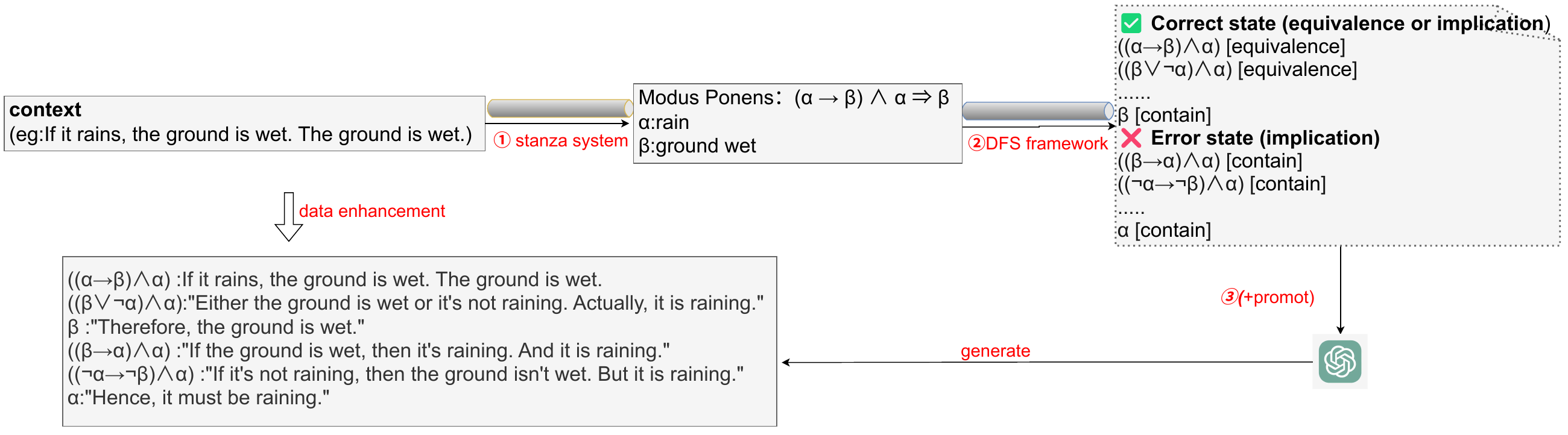}}
    \caption{Data Augmentation Pipeline: \textcircled{\scriptsize 1} Stanza System formalizes natural language into propositional logic formulas; \textcircled{\scriptsize 2} DFS Framework systematically explores and generates diversified novel formulas; building on this, \textcircled{\scriptsize 3} Prompt drives the large language model to instantiate these new formulas into natural text. Ultimately, the pipeline Generates high-quality data to achieve Data Enhancement.}
    \label{fig:framework}
\end{figure*}

\section{Related Work}

Based on the varying degrees of contextual dependency and logical depth required by different tasks, natural language logical reasoning evaluation tasks can be primarily categorized into two types: sentence-level reasoning, which focuses on examining immediate logical relationships within or between sentences, such as the natural language inference benchmarks MNLI~\citep{williams2018broad} and MRPC~\citep{dolan2005automatically}; and the more challenging passage-level reasoning, which requires models to integrate and deduce information from long-text contexts, exemplified by multiple-choice question-answering datasets like ReClor~\citep{yu2020reclor} and LogiQA~\citep{liu2020logiqa}. Consequently, to enhance the logical reasoning capabilities of models, several primary approaches have been pursued:

Path 1: At the data level, the core of this approach lies in injecting formal logical structures into training data through external tools or rules, enabling models to implicitly learn logical rules. Its effectiveness is built upon strict adherence to formal logic theorems, thereby ensuring the reliability of the generated data. The evolution of this method demonstrates a clear trajectory: AMR-DA\citep{shou2022amr} uses Abstract Meaning Representation as a bridge to enhance structured semantic associations through AMR graph reconstruction, laying the methodological foundation; LReasoner\citep{wang2021logic} leverages syntactic analysis and employs logical templates to generate equivalent sentences, explicitly introducing symbolic rules; AMR-LDA\citep{bao2023abstract}, building on this, directly applies broad-spectrum logical equivalence laws on AMR graphs for precise transformations, achieving more systematic logical data augmentation.

Path 2: At the model level, this approach focuses on directly eliciting or reshaping the model's inherent reasoning capabilities, operating under the premise that it already possesses substantial reasoning abilities (e.g., GPT-3.5\citep{achiam2023gpt}), It does not rely on external symbolic systems but directly activates its inherent potential through prompt engineering or training strategies, Chain-of-Thought\citep{wei2022chain} enhances multi-step reasoning by explicitly generating intermediate reasoning steps, while Self-Consistency \citep{wang2022self-2} improves output robustness by aggregating results from multiple reasoning paths.

Path 3: Enhancing the foundational logical reasoning capabilities of models at the parameter level. For instance, MERIt~\cite{jiao2022merit} employs a meta-path strategy to identify logical text structures, while IDoL~\cite{xu2023idol} constructs pre-training datasets using logical indicators to enhance the model's logical reasoning abilities.

In this context, we propose the LFC-DA method, which falls under Path 1 described above. Compared to LReasoner, it demonstrates superior systematicity in variant generation and produces more natural textual expressions. When contrasted with AMR-LDA, which relies on limited graph transformations, the core mechanism of LFC-DA enables more comprehensive coverage of diverse logical equivalences and entailment forms through systematic exploration of the logical space.

\section{method}
\subsection{system framework}
In this section, we introduce LFC-DA (Logical Formula-Controlled Data Augmentation), an innovative framework for logical data augmentation by establishing a formal pipeline, with its complete system architecture illustrated in Figure~\ref{fig:framework}. Our method comprises three coherent stages: Formalization, where the framework analyzes input texts through sophisticated linguistic processing to extract logical structures and converts them into \textbf{propositional logic formulas} (\S~\ref{sec:formalization}); Exploration, which systematically explores the logical equivalence space using state-space search algorithms to generate \textbf{diverse logical variants} (\S~\ref{sec:exploration}); and Instantiation, where the framework instantiates these logical forms into natural language expressions through LLM-based generation, producing rigorously controlled training data to enhance the \textbf{reasoning performance} of pre-trained models (\S~\ref{sec:instantiation}).
%3.2
\subsection{Stanza logical analysis framework}
\label{sec:formalization}
\begin{figure*}[htbp]
    \centering
    \includegraphics[width=0.9\textwidth]{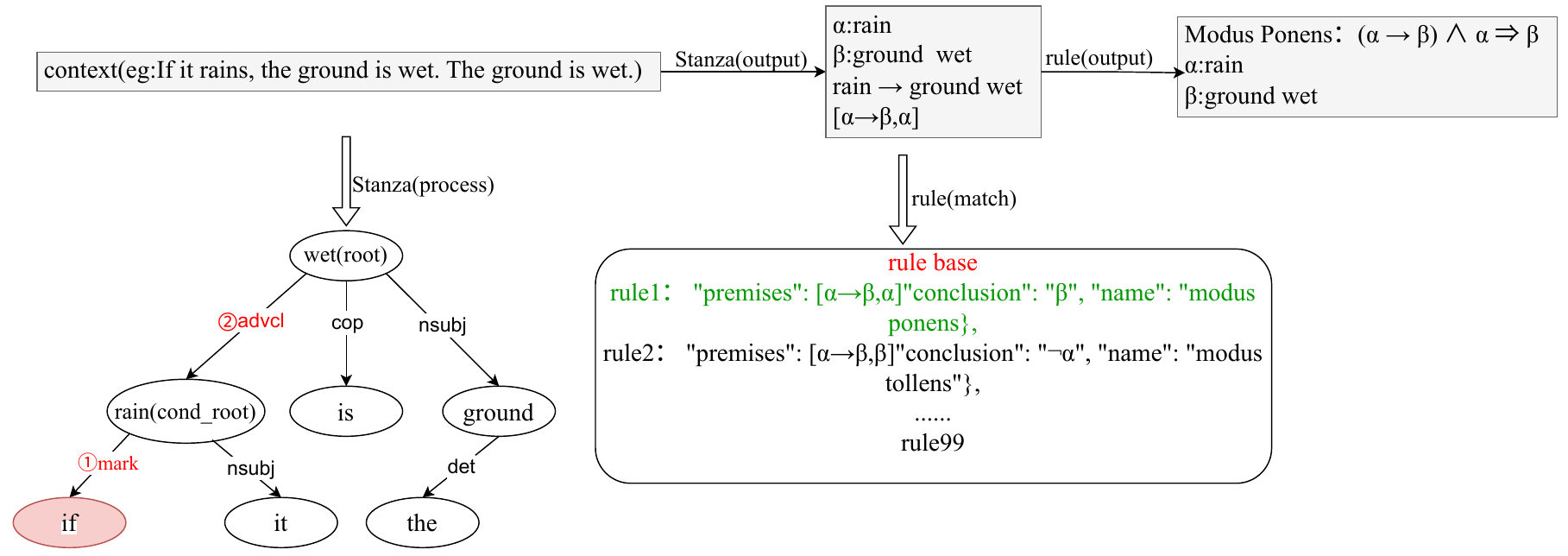}
    \caption{
    This figure illustrates the conversion pipeline from natural language instances to propositional logic formulas: 
    \textbf{Stanza(process)} performs syntactic parsing; 
    \textbf{Stanza(output)} identifies the logical structure $\alpha \rightarrow \beta$ and defines variables ($\alpha$: rain, $\beta$: ground wet); 
    \textbf{rule(match)} matches $[\alpha\rightarrow\beta, \alpha]$ with the rule base, after which 
    \textbf{rule(output)} generates the final formula $(\alpha\rightarrow\beta) \wedge \alpha \Rightarrow \beta$.
    }
    \label{fig:stanza-pipeline}
\end{figure*}
This paper introduces a three-stage pipeline for converting logical natural language into propositional logic representations. The framework operates as follows: (1) Syntactic Parsing: employing Stanza to parse input sentences into syntactic trees, providing a structural foundation for subsequent analysis\citep{qi2020stanza};(2)Logical Focus Identification: pinpointing logical operators and their scopes within the parsed structure to extract core logical information; (3) Formalization: abstracting the identified logical elements into well-formed propositional logic formulas. This pipeline establishes a robust correspondence between natural language expressions and formal logic rules.\\
\noindent\textbf{Syntactic Structure Parsing and Focus Identification:} Logical focus identification is built upon deep syntactic structure parsing, as shown in Figure 4. First, a pre-trained Stanza model is employed to perform grammatical parsing on input sentences (\texttt{Stanza(process)}), converting linear text into a hierarchical syntactic tree (including \textbf{part-of-speech (POS) tagging}, \textbf{lemmatization}, and \textbf{dependency relations}), providing a structured foundation for logical analysis. On this basis, focus identification adopts a collaborative strategy of ``\textbf{explicit triggers first, dependency relations supplemented}'': \textcircled{1} The \textbf{explicit trigger method} locates logical components and eliminates ambiguities (e.g., ``if'' in object clauses) through dictionary matching (e.g., conditionals like ``if'', connectives like ``and'') and verification of syntactic roles (e.g., the \texttt{marker} label); \textcircled{2} The \textbf{dependency relation method} captures implicit logical relationships by analyzing specific grammatical labels (e.g., \texttt{advcl} for conditionals, \texttt{conj} and \texttt{cc} for coordination). The two methods work collaboratively to precisely determine the type and scope of logical operators.
\\\textbf{Logical Rule Construction and Abstraction Stage:} Based on the results of logical focus identification (\texttt{Stanza(output)}), the following steps are executed sequentially: component extraction and phrase purification, filtering function words and constructing lemmatized phrases; propositional variable mapping, ensuring a unique correspondence between phrases and variables through a global mapping table, and deriving conclusions based on the rule base (\texttt{rule(match)}); logical formula generation, constructing standardized formulas based on logical operators and propositional variables (\texttt{rule(output)}).

%3.3
\subsection{DFS enumeration of logical deduction}
\label{sec:exploration}
Based on the aforementioned logical rules, this paper constructs a depth-first search (DFS) based branch enumeration framework to achieve systematic transformation of propositional logic formulas and exploration of the state space. The core operational logic of the framework can be abstracted as a cycle of "\textit{rule-driven rewriting}—\textit{pattern matching verification}—\textit{duplicate result pruning}—\textit{depth-first expansion}". This is realized through a quadruple technical design encompassing traversal strategy, rule definition, matching execution, and label verification, ultimately forming a complete transformation graph from basic formulas to their semantically equivalent forms.

\begin{figure}[!ht]  
    \vspace{-0.7em}
    \centering
    \includegraphics[width=0.4\textwidth]{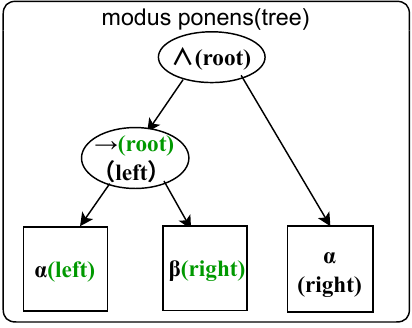}
    \caption{DFS enumeration process: Using the abstract syntax tree of \textit{Modus Ponens} as an example to illustrate the depth-first pre-order traversal strategy. The process starts from the (black) root node conjunction symbol $\wedge$, then systematically traverses the (green) left subtree (accessing its internal nodes in the order of $\rightarrow$, $\alpha$, $\beta$), and finally accesses the (black) right child node $\alpha$ under the root node.}
    \label{fig:dfs-enumeration}
\end{figure}

%3.3.1
\subsubsection{DFS Traversal and Pruning Mechanism}
The framework employs a depth-first pre-order traversal strategy, systematically scanning the abstract syntax tree in a fixed `root-left-right' sequence. As shown in Figure 2, processing begins at the root conjunction operator ($\wedge$), where rule matching is first attempted. The system then proceeds to the left subtree in a predetermined order: sequentially visiting the implication node ($\rightarrow$) and its left and right child nodes. After fully exploring the left subtree, the system continues to process the right subtree according to the predefined sequence. This deterministic traversal order ensures that logical rules can be systematically applied to each syntactic component in the AST, enabling organized exploration of the proof space.

During exploration, the system invokes applicable inference rules from the rule base (such as \textit{modus ponens} and conjunction commutativity). While the traversal strategy draws inspiration from the traditional backtracking search concept of "systematically exploring all possible branches", it achieves a stateless derivation mechanism through an immutable AST. As illustrated in Figure 4, once an AST node is created, its structure remains permanently fixed. Rule applications do not modify the original nodes but recursively generate entirely new successor subtrees. Consequently, parent context (including node structure, exploration progress, and path information) is naturally preserved in the upper frames of the recursive call stack, eliminating the need for explicit state saving and backtracking recovery steps in traditional methods. The depth-first search process can be formalized by the recursive expression shown in \eqref{eq:dfs-formula}.

\begin{equation}
\begin{split}
\mathrm{DFS}(s,d,P,V) =
\begin{cases} 
\{P\}, & \text{if } s = \text{target} \\[1pt]
\emptyset, & \text{if } d \geq d_{\max} \\[1pt]
\emptyset, & \text{if } R(s) \setminus V = \emptyset \\[1pt]
\bigcup_{\substack{s' \in R(s) \setminus V}} & \mathrm{DFS}(s',d+1, \\
& P \parallel s', \\
& V \cup \{s'\}), \quad \text{otherwise}
\end{cases}
\end{split}
\label{eq:dfs-formula}
\end{equation}

For the termination timing of invalid branches, the termination condition is formally defined, with specific triggering scenarios including depth exceeding limits ($d \geq$ preset upper bound), rule exhaustion (no new rules applicable), etc. When this condition is met, the framework terminates the exploration of the current branch, automatically reverts to the parent node context via the recursive call stack, and seamlessly transitions to the next parallel branch (i.e., the new subtree generated by previously untried rules from the parent node).

\begin{equation}
\smash{\mathrm{Backtrack}_{\text{Condition}}(s,d,V) = (R(s) \subseteq V) \vee (d \geq D_{\max})}
\end{equation}

%3.3.2
\subsubsection{Fault-tolerant verification mechanism}
This paper introduces error rules (label=0) into the rule set, designed according to the principle of ``syntactically valid but logically invalid'' (e.g., rule \textcircled{7} in the \textit{logic rule base} incorrectly deduces ``$Y \rightarrow X$'' from ``$X \rightarrow Y$'', which are logically non-equivalent). During framework execution, error rules and correct rules (label=1) share the same rule application and state management mechanism. However, they are accurately distinguished through a label propagation mechanism: the label of the current path is assigned the minimum value between the parent node's label and the current rule's label (correct\textcircled{6} + error\textcircled{4} = error), and erroneous branches are marked as label=0. Since multiple derivation paths may exist, to avoid redundancy, the complete information of each path is recorded. Subsequent occurrences of duplicate branches are directly pruned, retaining only unique instances.

The introduction of error rules serves three purposes: \textbf{first}, to verify the framework's ability to distinguish between valid and invalid paths; \textbf{second}, to simulate practical reasoning errors, enhancing the framework's practical validation; \textbf{third}, to provide natural positive and 

\begin{equation}\mathcal{L}=-\sum\log\frac{\exp\left(h\left(+\right)\right)}{\exp\left(h\left(+\right)\right)+\exp\left(h\left(-\right)\right)},\end{equation}

%3.4
\subsection{Display the reasoning steps}
\begin{figure*}[!t]
    \centering
    \fbox{\includegraphics[width=0.88\textwidth]{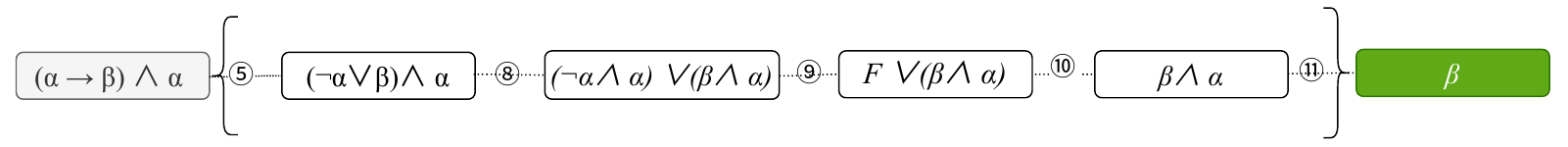}}  
    \caption{Alternative Derivation Path: This figure illustrates an alternative derivation path from the premise $(\alpha \rightarrow \beta) \wedge \alpha$ to the conclusion $\beta$. Assuming rule \textcircled{1} is unavailable in the rule base, the system systematically applies fundamental theorems from the rule base, progressively deriving the target through a series of rigorous logical transformations.}
    \label{fig:dfs-enumeration-1}
    
    \vspace{0.8cm}
    
    \centering
    \fbox{\includegraphics[width=0.88\textwidth]{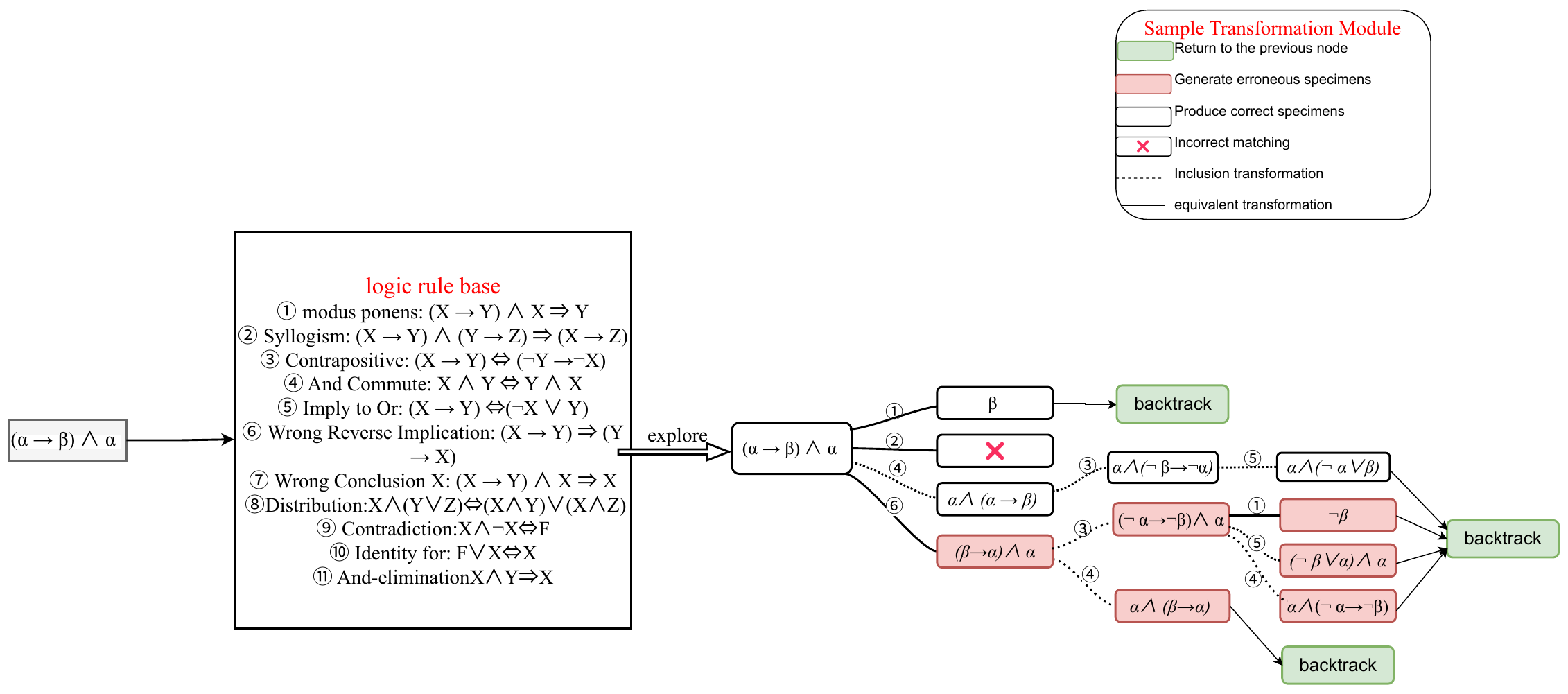}} 
    \caption{Symbol-Text Bimodal Data Generation Framework: This framework integrates symbolic logic with LLM text generation capabilities to construct interpretable training data. The system first expands the initial formula into correct state set (C) and error state set (E) based on the rule base. Subsequently, based on three logical relations (\textcircled{1} equivalence, \textcircled{2} implication, \textcircled{3} combination derivation), it systematically constructs labeled training sample pairs (Label 1 for positive examples, Label 0 for negative examples). Finally, through customized Prompt, the large language model is driven to instantiate symbolic samples into natural text.}
    \label{fig:bimodal-framework}
\end{figure*}

To achieve transparency and traceability in logical reasoning, we draw inspiration from the search methodology of Prolog. Since each valid input-output logical rule pairing is accompanied by complete reasoning steps, we construct a reasoning step display mechanism by integrating our depth-first search framework with the rule application system. The core objective of this mechanism is to take the input logical formula as the premise, preset the target formula as the conclusion, and record then display the complete transformation trajectory from start to finish. This provides an intuitive derivation path for understanding logical relationships between formulas. Upon successfully reaching the target formula, the system organizes and outputs the reasoning steps recorded during the search process.

For transformations that affect the entire formula, the system generates a concise expression in the format \texttt{'RuleName: OriginalExpr → NewExpr'}. For substitutions involving subformulas, it displays a contextual description formatted as \texttt{'RuleName: SubExpr → NewSubExpr within ParentExpr'}, ensuring the derivation process remains clear and readable. The derivation steps recorded by this mechanism enable visual verification of logical transformations. By incorporating such derivation records with explicit reasoning paths, we establish a reference paradigm for constructing logical reasoning training datasets, demonstrating fundamental patterns of rule-based derivation processes.

\subsection{Logical enhancement of data generation}
\label{sec:instantiation}

\begin{figure*}[htbp]
    \centering
    \includegraphics[width=0.9\textwidth]{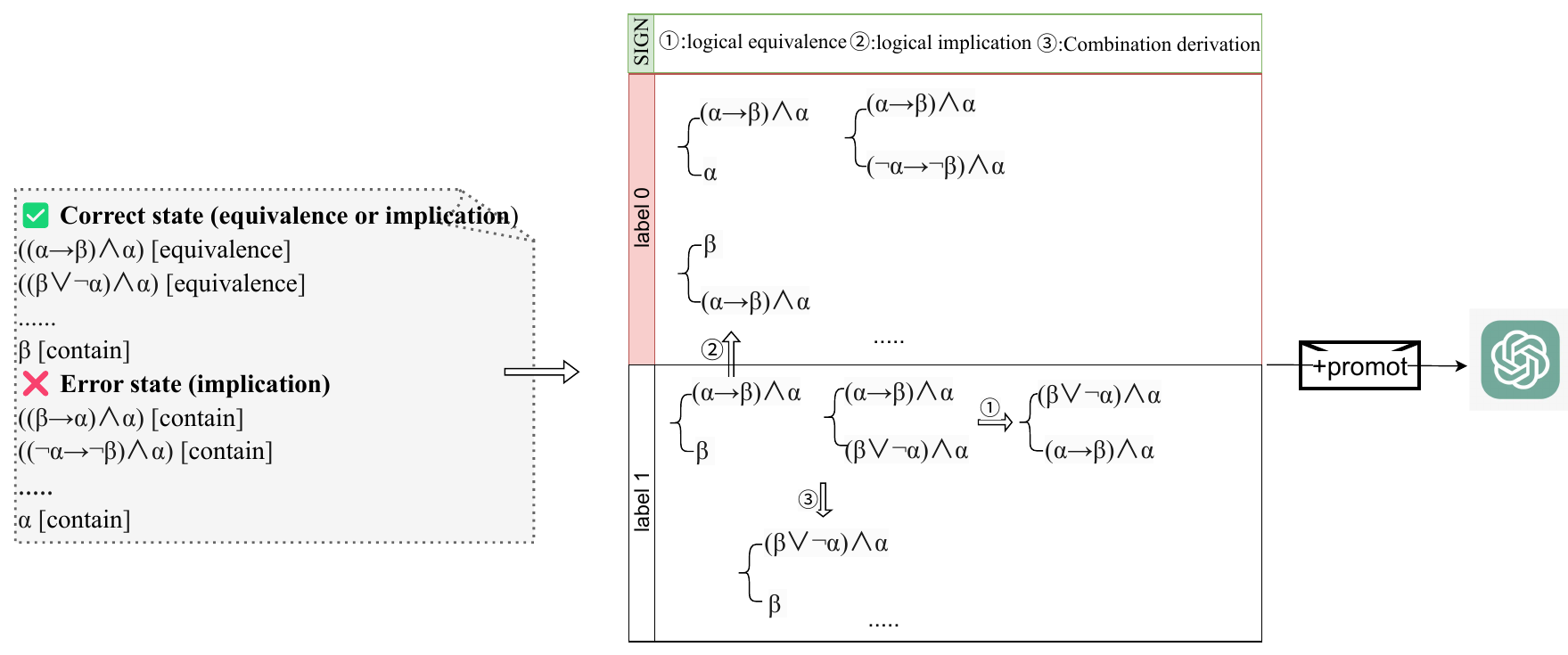}
    \caption{This framework integrates symbolic logic with LLM capabilities to create interpretable training data. It expands initial formulas into correct (\textbf{OK}) and error (\textbf{ERROR}) state sets, annotates logical attributes via [equivalence] and [contain] markers, constructs labeled sample pairs using three logical relations (\textcircled{1} equivalence, \textcircled{2} implication, \textcircled{3} derivation), and employs customized prompts to instantiate symbolic samples as natural text, ensuring logical correctness, diversity, and interpretability.}
    \label{fig:stanza-pipeline}
\end{figure*}
Through a symbol-text bimodal driven approach, we leverage the powerful text generation capabilities of large language models (LLMs) to achieve accurate conversion from logical symbols to natural language, while ensuring explainability via a structured design. Specifically, the textual modality provides concrete semantics for the symbols. Combined with customized prompts, it guides the LLM to generate natural language that is both logically rigorous and semantically fluent, avoiding unsubstantiated free fabrication, generation hallucinations\citep{gao2023pal}, and deviations from the original logic.

Explainability is realized through a two-layer design: first, the generated results can be directly traced back to the source symbolic logic, ensuring clear provenance; second, the logical states are divided into a correct state set \(S_1\) (logically self-consistent and semantically coherent) and an erroneous state set \(S_2\) (containing logical contradictions or semantic conflicts). We systematically construct two types of pairs: ``\(S_1\)-\(S_1\) (label=1, positive examples)'' and ``\(S_1\)-\(S_2\) (label=0, contrastive examples)''. This construction method is based on the fundamental differences between logical implication and equivalence relations: implication relations are unidirectional (corresponding to \textcircled{2} in Figure 1), and swapping the premise and conclusion yields invalid reasoning, automatically forming contrastive negative examples; whereas equivalence relations are bidirectionally symmetric (corresponding to \textcircled{1} in Figure 1), and swapping the order remains valid reasoning, thus allowing the construction of bidirectional positive examples to ensure logical rigor. Furthermore, through combined derivation (corresponding to \textcircled{3} in Figure 1), equivalence and implication relations are \textbf{linked} to form multi-step reasoning chains, thereby constructing training samples that cover more complete reasoning processes.
%4
\section{Experiment}
%4.1
\subsection{Public dataset}
In terms of model evaluation, this paper selects two challenging logical reasoning datasets, ReClor and LogiQA. ReClor is derived from GMAT and LSAT exam questions, focusing on logical analysis and deductive reasoning of complex texts. LogiQA, on the other hand, comes from civil service exams and emphasizes logical judgment and inductive reasoning in real-world contexts. In addition, the datasets for MNLI (Natural Language Inference, NLI) and MRPC (Microsoft Research Paraphrase Corpus) tasks are also chosen. These four datasets are used together for comprehensive evaluation after the second-stage fine-tuning of the models.
For the first-stage training data, this paper selects 9 classic statements converted into logical symbol forms. Training samples are constructed by generating both correct and incorrect state sets. Given that the number of transformation rules is large, we extract key form variants (including implication and equivalence) for each rule. Then, using large models, we generate 7 different stylistic templates for each logical form.
%4.2
\subsection{experimental facilities}
The experiments were conducted on a server equipped with 2 NVIDIA GeForce RTX 3090 GPUs, each with 24GB of VRAM. The code is implemented using the PyTorch 1.12 framework and Python 3.9, with acceleration provided by CUDA 11.6. The key libraries used include Transformers 4.20.0, Datasets 2.0.0, and the pre-trained model RoBERTa-large. For generating training data, large models GPT-4 and DeepSeek were employed. To ensure the reproducibility of the results, all experiments were conducted with a fixed random seed (42).

%4.3
\subsection{Generate contrastive learning data for training discriminative models}
\begin{table}[!t]
\centering
\small 
\setlength{\tabcolsep}{4pt} 
\begin{tabular}{lcccc}
\toprule
\multirow{2}{*}{\textbf{Models/ Datasets}} & 
\multicolumn{2}{c}{\textbf{ReClor}} & 
\multicolumn{2}{c}{\textbf{LogiQA}} \\
\cmidrule(r){2-3} \cmidrule(l){4-5}
 & \textbf{Dev} & \textbf{Test} & \textbf{Dev} & \textbf{Test} \\
\midrule
RoBERTa & 61.40 & 54.10 & 34.72 & 35.35 \\
RoBERTa LReasoner-LDA & 62.70 & 57.9 & 39.01 & 36.7\\
RoBERTa AMR-LDA & 63.53 & 56.70 & 38.5 & 37.8 \\
LFC-DA & \textbf{64.77} & \textbf{59.70} & \textbf{39.17} & \textbf{40.1} \\
\bottomrule
\end{tabular}
\caption{Model Performance Comparison on Logical Reasoning Datasets}
\label{tab:model_comparison}
\end{table}
This section aims to verify whether contrastive learning with synthetic data generated using the LFC-DA method can improve the performance of discriminative language models on downstream logical reasoning tasks. We use AMR-LDA and LReasoner-LDA as baselines and perform contrastive experiments under fully unified conditions for second-stage fine-tuning (ReClor and LogiQA logical reasoning data).

The synthetic data for the first stage is independently generated by both parties, with approximately 8000 pairs for the training set and 3000 pairs each for the validation and test sets, maintaining a 1:1 positive-to-negative sample ratio. The key differences between methods lie in the logical rule sets and stylistic templates, which are not unified. The base model used is RoBERTa-Large, and after contrastive learning, all models are fine-tuned on the same downstream tasks. Results are evaluated based on "test set accuracy" as the core metric. LFC-DA significantly improves performance on most logical reasoning tasks, confirming its effectiveness, as shown in Table 2-1.

For the ReClor test set, the results are further categorized into "memory/matching" models and "true reasoning" models, corresponding to TEST-E and TEST-H. A comparison of these models is provided in Table 2-2.

\begin{table}[!t] 
\centering
\setlength{\tabcolsep}{8pt}
\begin{tabular}{lcc}
\toprule
\textbf{Model} & \textbf{Test-E} & \textbf{Test-H} \\
\midrule
RoBERTa & 74.77 & 37.85 \\
+ LReasoner-LDA & 76.10 & 43.57 \\
+ AMR-LDA & 75.09 & 40.89 \\
+ LFC-DA & 75.45 & 47.14 \\
\bottomrule
\end{tabular}
\caption{Performance on ReClor Easy and Hard Subsets}
\label{tab:reclor_easy_hard}
\end{table}

\section{Conclusion}
The core of this study lies in the proposal and validation of the LFC-DA method, which aims to solve the problem of logical consistency in data augmentation tasks. This method constructs a reliable set of input logic templates for large language models, effectively constraining their generation process to ensure that augmented data remains logically consistent with the original data, thereby significantly reducing the randomness and factual bias in the generated results. Empirical results show that the LFC-DA method not only fully utilizes the powerful generation capabilities of large models, transforming them into efficient and reliable auxiliary tools, but also significantly reduces the reliance on manual intervention while ensuring the accuracy and logical fidelity of the augmented data. This study confirms the effectiveness of combining structured logic templates with large model generation, providing a reliable technical pathway for high-quality data augmentation.
\section{Restrictions}

In this study, the syntactic analysis module is designed to process only simple sentences, capable of extracting basic logical focus and generating corresponding propositional variables. This design is sufficient to provide reliable input for subsequent DFS logical inference. However, it still has limitations when dealing with complex sentence structures, such as multiple clauses, nested conditionals, or long-distance dependency structures. Therefore, the current method has certain limitations at the syntactic analysis level.

\section{Future work}
Future work can be pursued by enhancing syntactic analysis capabilities through the integration of stronger syntactic/semantic analysis tools, such as DRT-based or Boxer-based semantic parsers, to support the extraction of complex sentence structures, nested conditionals, and multi-branch logic, thereby improving the completeness of logical formula generation. Additionally, exploring the introduction of modal or probabilistic logic could expand the logical expression capabilities, enabling the system to handle information about possibility, necessity, and uncertainty in natural language, which would enhance reasoning abilities in real-world contexts and complex texts. Furthermore, combining deep learning with pre-trained models for validity judgment and ranking of logical formulas could optimize DFS inference path selection, improving reasoning efficiency and accuracy. Finally, automating the discovery and learning of new logical rules by analyzing positive and negative samples in the inference process could reduce manual rule maintenance costs and enhance the system’s ability to adapt to different text types.

\bibliography{LFC-DA}

\end{document}